\documentclass[journal]{IEEEtran}
\usepackage{graphicx, subfigure}
\usepackage{enumerate}
\usepackage{amsmath,amssymb} 
\usepackage{color}
\usepackage{changes}

\usepackage{cite}

\ifCLASSINFOpdf
\else
\fi
\usepackage{amsmath}
\usepackage{array}

\ifCLASSOPTIONcompsoc
\else
\fi
\usepackage{url}

\begin{document}
%
\title{Video Trajectory Classification and Anomaly Detection Using Hybrid CNN-VAE}
%
%
%
\author{Kelathodi~Kumaran~Santhosh,~\IEEEmembership{Student Member,~IEEE,}
        Debi~Prosad~Dogra,~\IEEEmembership{Member,~IEEE,}
        Partha~Pratim~Roy,       
        and~Adway~Mitra
\thanks{K.~K.~Santhosh, D.~P.~Dogra and A. Mitra are with School of Electrical Sciences, Indian Institute of Technology Bhubaneswar, Odisha,
India e-mail: (sk47@iitbbs.ac.in, dpdogra@iitbbs.ac.in, adway@iitbbs.ac.in).}
\thanks{P.~P.~Roy is with the Department of Computer Science and Engineering, Indian Institute of Technology, Roorkee, India. e-mail:(proy.fcs@iitr.ac.in).}
}

%
%

\markboth{Pre-Print}%
{Shell \MakeLowercase{\textit{et al.}}: Bare Demo of IEEEtran.cls for IEEE Journals}
%



\maketitle

\begin{abstract}
   Classifying time series data using neural networks is a challenging problem when the length of the data varies. Video object trajectories, which are key to many of the visual surveillance applications, are often found to be of varying length. If such trajectories are used to understand the behavior (normal or anomalous) of moving objects, they need to be represented correctly. In this paper, we propose video object trajectory classification and anomaly detection using a hybrid Convolutional Neural Network (CNN) and Variational Autoencoder (VAE) architecture. First, we introduce a high level representation of object trajectories using color gradient form. In the next stage, a semi-supervised way to annotate moving object trajectories extracted using Temporal Unknown Incremental Clustering (TUIC), has been applied for trajectory class labeling. Anomalous trajectories are separated using t-Distributed Stochastic Neighbor Embedding (t-SNE). Finally, a hybrid CNN-VAE architecture has been used for trajectory classification and anomaly detection. The results obtained using publicly available surveillance video datasets reveal that the proposed method can successfully identify some of the important traffic anomalies such as vehicles not following lane driving, sudden speed variations, abrupt termination of vehicle movement, and vehicles moving in wrong directions. The proposed method is able to detect above anomalies at higher accuracy as compared to existing anomaly detection methods.
\end{abstract}


%
\IEEEpeerreviewmaketitle

\section{Introduction}
Timely detection of traffic anomaly is one of the prerequisites of an Intelligent  Transportation Systems (ITS). If not done timely, anomalies may create cascading effects leading to chaos in traffic. Typical examples of traffic anomalies are, lane driving violation, over-speeding, collision, red-light violation, etc. Anomaly detection using video object trajectories with deep learning has not yet been explored much. In this paper, we propose a color gradient approach for representing vehicular trajectories extracted from videos. These trajectories are then used for classification and anomaly detection at traffic junctions using a hybrid CNN-VAE architecture.

 Most commonly used features for video guided scene understanding are trajectories. A trajectory is a time series data with object locations indexed in temporal order. Classifying trajectories using neural networks is not trivial due to variation in the data length.  Key to the success of a time series signal classification lies in finding an effective representation of the data. Neural networks-based classifiers need fixed size inputs. CNN, Long Short Term Memory (LSTM) and Recurrent Neural Network (RNN) have been used for time series classification~\cite{yang2015deep, AKhosroshahi, hammerla2016deep}. However, time series data can be of varying length. . Therefore, classification of varying length data can be applied after preprocessing, e.g.  converting them into fixed length data either by padding or subsampling. If the trajectory length variance is large, preprocessing in mandatory. 

Video anomaly detection at traffic junctions is highly challenging due to its contextual nature. For example, when a signal turns green at a traffic junction, only a few of the paths or directions  are allowed for vehicle movement. Any motion that violates direction, is assumed to be anomaly though such motions can be normal in a different context. 
 
\begin{figure*}[h]
\centering
{\includegraphics[scale=0.24]{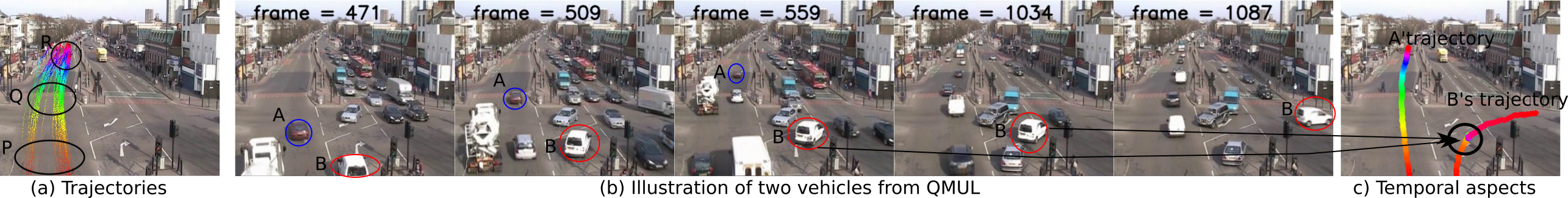}
}
\caption{Depiction of temporal characteristics using color gradient forms. (a) Trajectories from QMUL~\cite{QMUL_DATASET} junction dataset using color gradient representation. Temporal characteristics are similar at similar locations, marked as P, Q, and R. (b) Illustration of trajectory of a vehicle ($A$) that does not stop and another vehicle ($B$) that stops  and proceeds. (c) Temporal characteristics are significantly different for these two objects.}
\label{fig_MOTIVATION}
\end{figure*}

\subsection{Related Work}
\label{sec:Related_Work}
 Traditional features such as basis transform coding using wavelet and Fourier coefficients~\cite{bulling2014tutorial}, time series mean and covariance~\cite{bulling2014tutorial}, and symbolic representation~\cite{lin2003symbolic} have been used for classification of time-series data using neural networks. Also, other models such as  Deep Belief Networks (DBN)~\cite{tieleman2008training} have been used for human activity detection~\cite{plotz2011feature}. On the other hand, CNNs are primarily used in image classification~\cite{krizhevsky2012imagenet, zeiler2014visualizing}, activity recognition in videos~\cite{ji20133d}, speech recognition~\cite{deng2013recent}, etc. 
 
  Long  Short  Term  Memory  networks (LSTMs)~\cite{hochreiter1997long} are a special kind of Recurrent Neural Network (RNN) that can be used for handling sequential/time series data. Authors of~\cite{ng2015beyond, donahue2015long} have proposed a recurrent network connecting LSTMs to CNNs  to perform action recognition and video classification, respectively. Donahue et al.~\cite{donahue2015long} have tested the learned models for activity recognition, image description and video description. The work proposed in~\cite{wu2015modeling} has achieved the state-of-the-art performance in video classification by connecting CNNs and LSTMs under a hybrid deep learning framework. Sequential Deep Trajectory Descriptor (DTD) has been used for action recognition~\cite{shi2017sequential} from the video sequences. Deep Neural Network (DNN)-based trajectory classification has been applied on Global Positioning System (GPS) trajectories~\cite{Endo2016}. Dense feature trajectories used have been utilized for action recognition in videos ~\cite{HWang}. The LSTM-based work proposed in~\cite{AKhosroshahi} uses fixed size features to classify trajectories of surrounding vehicles at four way intersections based on LIDAR (LIght Detection And Ranging), GPS, and inertial measurement unit (IMU) measurements. 
  
  Dense trajectories extracted using neural networks have also been used for action recognition in videos including classifying a person when walking, running, jumping~\cite{HWang, 2001_AF_Bobick}, etc. These methods cannot handle multiple actions present in a scene. However, in real life scenario,  multiple objects can interact resulting more than one action within the scene. Training neural networks for action recognition can be challenging in presence of multiple activities.  However, object trajectories extracted using traditional methods~\cite{benfold2011stable, AMilan1, SHBae} can be used for learning the motion patterns using DNNs as they can automatically extract features from trajectories. The trained/learned model can then be used in classification and action recognition applications. 
  
In this work, we encode video trajectories using a high-level representation, named color gradient, that embeds spatio-temporal
information of the objects-in-motion. The high-level representation is then used for trajectory classification and anomaly detection using a hybrid CNN-VAE architecture. 

\subsection{Motivation and Contributions}
\label{sec:Motivation}
Since accurate classification is the key to detect anomalies, a classifier that can handle time series data with length variations, has been preferred. Typical neural networks-based methods need fixed input size. Therefore, varying length trajectories cannot directly be used in such classifiers. Conventional methods such as the one proposed in~\cite{yang2015deep} convert the varying length time series data into fixed size by sampling. This is similar to quantization, which leads to information loss. The question is: Why can't a trajectory represented using an image be given as an input to a classifier? However, trajectories representing movement of more than one object in between two locations may look visually similar when projected in 2D space. Such representations fail to preserve temporal relations between successive points of a trajectory.  Encoding of time information in the form of color gradient (red $\to$ violet)  reveals, similar patterns produce similar color gradient as depicted in Fig.\ref{fig_MOTIVATION}(a).  Similarly, the trajectories with possible anomalies exhibit different spatio-temporal characteristics as depicted in Fig.\ref{fig_MOTIVATION}(b). This has motivated us to propose the following: 

\begin{enumerate}[(i)]
\item A high-level representation of object trajectories using color gradient that encodes spatio-temporal information of trajectories of varying length.
\item A semi-supervised labeling technique based on modified Dirichlet Process Mixture Model (mDPMM)~\cite{TUIC_ITS_2018} clustering to identify the trajectory classes.
\item A method using t-Distributed Stochastic Neighbor Embedding (t-SNE)~\cite{maaten2008visualizing} to eliminate anomalous trajectories in the training data. 
\item Detection of traffic anomalies using a hybrid CNN-VAE architecture.  
\end{enumerate}
Rest of the paper is organized as follows. In Section~\ref{sec:Methodology}, we present the proposed methodology. Section~\ref{sec:Experimental_Results} presents experimental results and Section~\ref{sec:Conclusions} presents conclusion.

\section{Methodology}
\label{sec:Methodology}

First we discuss the background of the terms and concepts used in the work. A scene represents the  view captured using static camera. We use observation or data to represent a trajectory.  A cluster is a collection of trajectories of similar characteristics. A class is a set of trajectories having some selected common characteristics. Here, a class typically represents a unique path in a scene. A model is a representation of a real-world phenomenon. Here, model represents the weight parameters of the trained neural networks. We assume a model can represent a scene. Reconstruction loss (of CNN-VAE architecture) represents a measure of deviation from the input. A typical anomaly represents deviation from the normal path. Some anomalies are known a-priori. For example, when a signal turns green at a traffic junction, only a few of the paths are allowed for vehicle motion. Any motion that conflicts/intersects the allowed path, is considered as known anomaly. However, some anomalies may not be present in the training data. We refer them to as unknown anomalies.

\begin{figure*}[h]
\centering
{\includegraphics[width=0.9\textwidth]{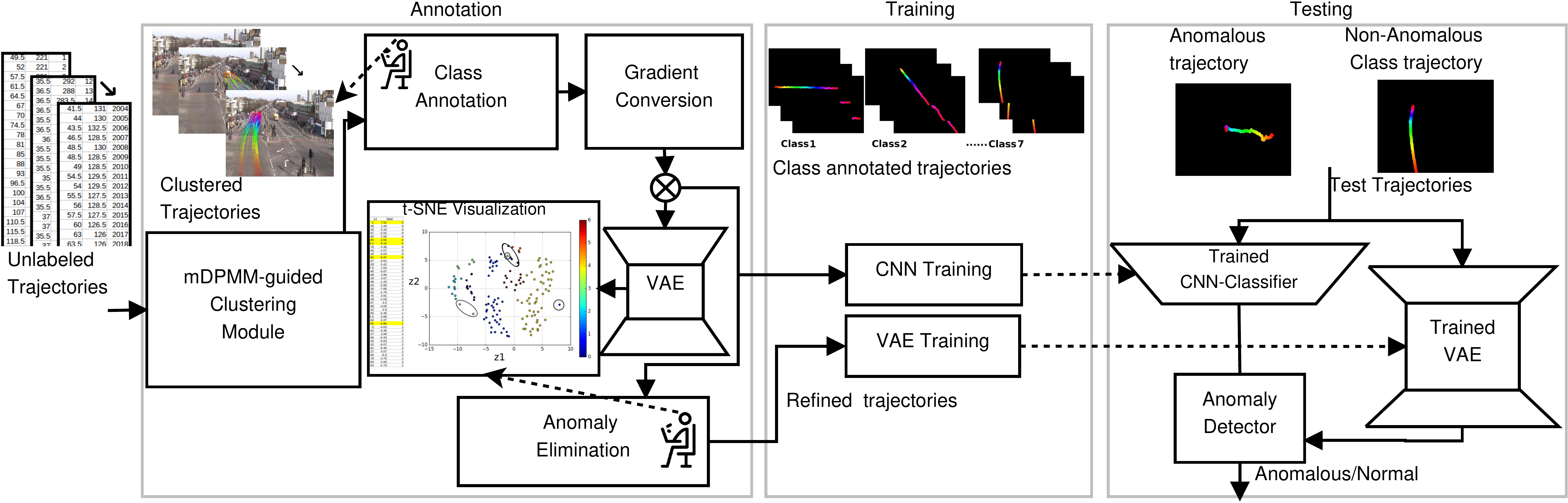}
}
\caption{Proposed anomaly detection framework. Unlabeled trajectories are grouped using the modified Dirichlet Process Mixture Model (mDPMM)~\cite{TUIC_ITS_2018}. Clusters are then mapped to different classes of trajectories using manual annotation and gradient representation is produced. These trajectories are then fed to train CNN and VAE to get t-SNE for eliminating anomalous trajectories (if any). The refined normal trajectories represented in color gradient form are fed to VAE to train the anomaly model of VAE. Anomaly detection is done once average reconstruction loss is known for the trained VAE with normal trajectories. Trained models are then used for classification and anomaly detection. }
\label{fig_anomaly_framework}
\end{figure*}  
 Object trajectories are obtained using~\cite{SHBae, TUIC_ITS_2018}. A trajectory ($\tau_i$) can be represented using~(\ref{equation:TRAJ}), where $(x_{l},y_{l})$ represents the position of moving object at time $t_l$ and $L_i$ be its length.  A cluster is a collection of trajectories of similar characteristics. A class is a set of trajectories having some selected common characteristics. It can be trajectories in the same lanes, trajectories following same route, etc. 
\begin{equation}
\tau_i  =  <(x_{0}, y_{0},t_{0}),(x_{1},y_{1},t_{1}), \cdots, (x_{L_i},y_{L_i},t_{L_i})>
\label{equation:TRAJ}
\end{equation}

 Traffic anomalies can be classified into two types; \textit{known} and \textit{unknown}.  \textit{Known} anomalies correspond to trajectories that may be allowed in different contexts. On the contrary, \textit{unknown} anomalies correspond to trajectories that are not present in the training data. In order to detect both types of anomalies, it is important to learn the normal trajectory patterns or classes. The overall anomaly detection framework is presented in Fig.~\ref{fig_anomaly_framework}. 
\subsection{Background} 
\subsubsection{Modified DPMM Guided Clustering}
When raw trajectories are obtained from some tracking algorithms, they need  to be clustered to identify different patterns.  
In~\cite{TUIC_ITS_2018}, we have proposed a modified DPMM (mDPMM) to group pixels having similar characteristics. Here, we use mDPMM to group trajectories to learn the motion patterns. The model is expressed using~(\ref{equation:DPM1}~-~\ref{equation:DPM4}).
\begin{equation}
z_i|\pi  \sim  \mbox{Discrete}(\pi)
\label{equation:DPM1}
\end{equation} 
\begin{equation}
\tau_i|z_i, \theta_k  \sim  F(\theta_{z_i})
\label{equation:DPM2}
\end{equation} 
\begin{equation}
\pi |e^{-\beta}  \sim  \mbox{Dirichlet}(e^{-\beta} / K, \cdots, e^{-\beta} / K)
\label{equation:DPM3}
\end{equation} 
\begin{equation}
\theta_k|H  \sim  H
\label{equation:DPM4}
\end{equation} 
Here, $\tau_i$ is a random variable representing the trajectory and $z_i$ corresponds to the latent variable representing cluster labels. $z_i$ takes one of the values from $k = 1 \cdots K$, where $K$ is the number of clusters. $\pi= (\pi_1, \cdots,\pi_K)$, referred to as mixing proportion, is a vector of length $K$ representing the probabilities of $z_i$ to be $k$. $\theta_k$ is the parameter of cluster $k$ and $F(\theta_{z_i})$ denotes the distribution defined by $\theta_{z_i}$. $e^{-\beta}$ is the concentration parameter of Dirichlet distribution and its value decides the number of clusters formed. $\beta$ is referred to as concentration radius. Trajectory clustering is to be done by taking $\tau_i$ as $<$$x_{s},y_{s},x_{e},y_{e},t_{d}$$>$, where $(x_{s}$, $y_{s})$ represents the start position, $(x_{e},y_{e})$ the end position and $t_{d}$ is the duration/length of the trajectory.

	Using the inference method given in~\cite{TUIC_ITS_2018}, clustering of clustering of trajectories can be done. These clusters can be typically grouped into two types. First type contains large number of trajectories and they represent prominent patterns in the scene. The second type of clusters contain less number of trajectories. They can either correspond to less frequently occurring patterns or anomalies.
\subsubsection{Gradient Conversion of the Trajectories}
A trajectory in time series is mapped into a color gradient form by varying hue using $ hue(x_{l},y_{l}) = (t_l - t_0)/L_i*180$, $0\leq l \leq L_i$ within an image frame. These gradient frames become inputs to the CNN and VAE.
\subsubsection{Anomaly Elimination in Training Data using t-SNE}
 t-SNE~\cite{maaten2008visualizing} is a machine learning algorithm for visualizing high-dimensional data in a low-dimensional space. We use this for visualizing latent features of a trained VAE in two dimensions. Trajectories belonging to same class typically lie in close proximity in the visualization plane. However, trajectories that are far away from a class are inspected again for manual anomaly checking.
\subsection{Trajectory Annotation}
Suppose a set of trajectories captured from a traffic junction or road are given. These trajectories must belong to any one of the defined set of paths (classes). Applying mDPMM helps to identify prominent patterns from these trajectories. Like any unsupervised method, clustering algorithm can only identify different possible patterns from the trajectory data. Though prominent patterns can correspond to normal trajectories, clusters with less number of trajectories can represent a rare pattern or an anomaly. This necessitates to have an additional annotation process to identify allowed classes. Clustering reduces the load of the manual labeling process as an initial grouping is done through mDPMM. The annotator can identify these rare patterns through visual observation of the scene and separate the anomalous trajectories to finalize the allowed classes. This process is called class annotation.

More refinements are possible within a class. It is possible that two trajectories with similar endpoints and duration may follow different paths, out of which one may be normal. This may not always be detected through visual observation. Therefore, t-SNE has been used to visualize the distribution of trajectories within the classes. This helps to remove noises (anomalies) from the training set being prepared for VAE.
\subsection{Training CNN and VAE Framework}
\begin{figure}[h]
\centering 
\subfigure[CNN Classifier]{\includegraphics[width=0.48\textwidth]{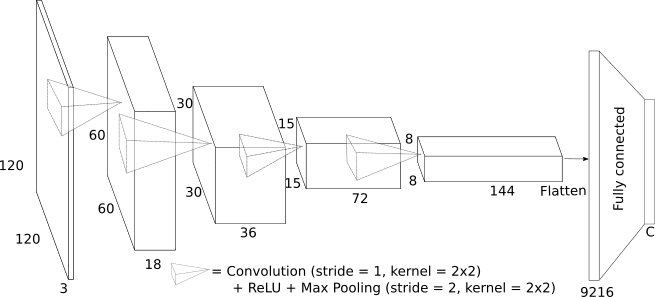}
\label{fig_single_trajectory}}
\hfill
\subfigure[Variational Autoencoder (VAE)]{\includegraphics[scale = 0.22]{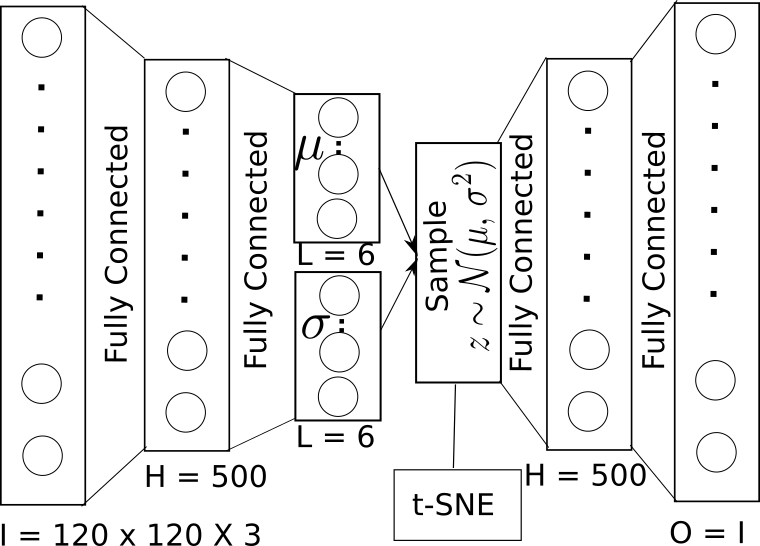}
\label{fig_single_trajectory}}
\caption{CNN and VAE  details. (a) Proposed CNN architecture for trajectory classification. (b) Autoencoder with the dimensions of each layer. $I/O$, $H$ and $L$ represent input/output, hidden and latent dimensions, respectively. }
\label{fig_CNN_AUTO}
\end{figure}

 A CNN classifier typically consists of repeated occurrences of cascaded convolution,  activation, and pooling layers followed by fully connected layers. The architecture used in this work is depicted in Fig.~\ref{fig_CNN_AUTO}(a). During the training stage, a cost/loss function representing the cross-entropy between the expected and predicted class is minimized using Adam optimizer~\cite{AdamOptimizer} with a learning rate of $\lambda$.    

We use a variational autoencoder (VAE) similar to~\cite{kingma2013auto} to detect unknown anomalies.  It typically consists of encoding and decoding stages. Input to the encoder $q​_{\theta}​​(z|\tau)$ is $\tau$. Output is a hidden/latent feature $z$, where $\theta$ represents weights and biases of encoder network. Decoder $p​_{\phi}​​(\tau|z)$ takes latent feature $z$ and regenerates $\tau$, where $\phi$ represents weights and biases of decoder. Loss function ($l_i$) for a trajectory $\tau_i$ is given in~(\ref{equation:vae_loss_function}) in terms of log likelihood ($ll$) as given in as given~(\ref{equation:vae_likelihood}) and Kullback-Leibler Divergence (KLD) as given in as given~(\ref{equation:KLD}). Adam optimizer minimizes the average loss function during training. Once trained, VAE can detect anomalies using the average reconstruction loss on the trained VAE. 
\begin{equation}
ll = \mathbb{E}_{z \sim q_{\theta}(z|\tau_i)}[log p_{\phi}(\tau_i|z)]
\label{equation:vae_likelihood}
\end{equation}
\begin{equation}
KLD = q_{\theta}(z|\tau_i)||p(z)
\label{equation:KLD}
\end{equation}
\begin{equation}
l_i(\theta,\phi) = -ll + KLD
\label{equation:vae_loss_function}
\end{equation}
\subsection{Anomaly Detection}
Classification is performed on the trained CNN using test trajectories represented in gradient form to obtain class $c$. Let $\delta$ be the threshold of reconstruction loss value for normal classes on the trained VAE. $\delta$ is derived using the variance of loss values on the training trajectories. A trajectory can be considered anomalous when $c \notin A_s$ or $l_i(\theta,\phi) > \delta$, such that $A_s$ is a set of allowed trajectory classes of a particular signal $s$. 

However, a classifier is needed for anomaly detection to handle conflicting trajectories. In a typical traffic junction, a set of flows may be allowed at a given time. For example, the QMUL dataset (Fig.~\ref{fig_QMUL_CLUSTERS}) suggests, any two flows, e.g, south-to-north on left side and north-to-south on right side, are allowed at a given time. Any other movements can be termed anomalous though individually such movements may be allowed at a different time. VAE cannot detect such known anomalies.  Therefore, CNN helps to detect such conflicting anomalies. It also helps to identify the anomalous path. 
\section{Experimental Results}
\label{sec:Experimental_Results}
We have used tensorflow and openCV for developing the classification and anomaly detection framework. We have used three datasets, namely T15\cite{xu2015unsupervised_t15}, QMUL~\cite{QMUL_DATASET} and a junction video dataset (referred to as 4WAY). Context tracker~\cite{ContextTracker} has been used for creating trajectories from QMUL dataset, and  Temporal Unknown Incremental Clustering (TUIC)~\cite{TUIC_ITS_2018} has been used for obtaining 4WAY trajectories. Inputs to CNN-VAE are resized to 120x120x3. CNN training has been completed with 50 epochs with a learning rate of $\lambda = 1e^{-3}$ on T15 and 4WAY dataset videos. A batch size of 20 has been used for the QMUL dataset. VAE for T15 has been trained with $\lambda = 5e^{-4}$ in 500 epochs using a batch size of 20. VAE for QMUL dataset has been trained with $\lambda = 1e^{-4}$ and batch size = 10 in 500 epochs.
\subsection{Experiments on Trajectory Clustering and Annotation}
\begin{table}[h]
\caption{Dataset annotation results. $T$ represents the video duration, $N$ the number of trajectories, $K$ the number of clusters obtained using mDPMM, $C$ the number of valid classes and $N_A$ the number of anomalous trajectories. T15 is a labeled dataset.}
\label{Tab:Dataset}
\begin{center}
\begin{tabular}{|l|r|r|r|r|r|r|}
\hline
Dataset & $\beta$& $T$ (min) & $N$ & $K$ & $C$ & $N_A$ \\
\hline\hline
T15 & - & - & 1500 & NA & 15 & 31\\
QMUL & 180 & 10 & 166 & 23 & 7 & 17\\
4WAY & 100 & 28 & 3861 & 193 & 18& 808\\
\hline
\end{tabular}
\end{center}
\end{table}
\begin{figure}[h]
\centering
\includegraphics[width=0.46\textwidth]{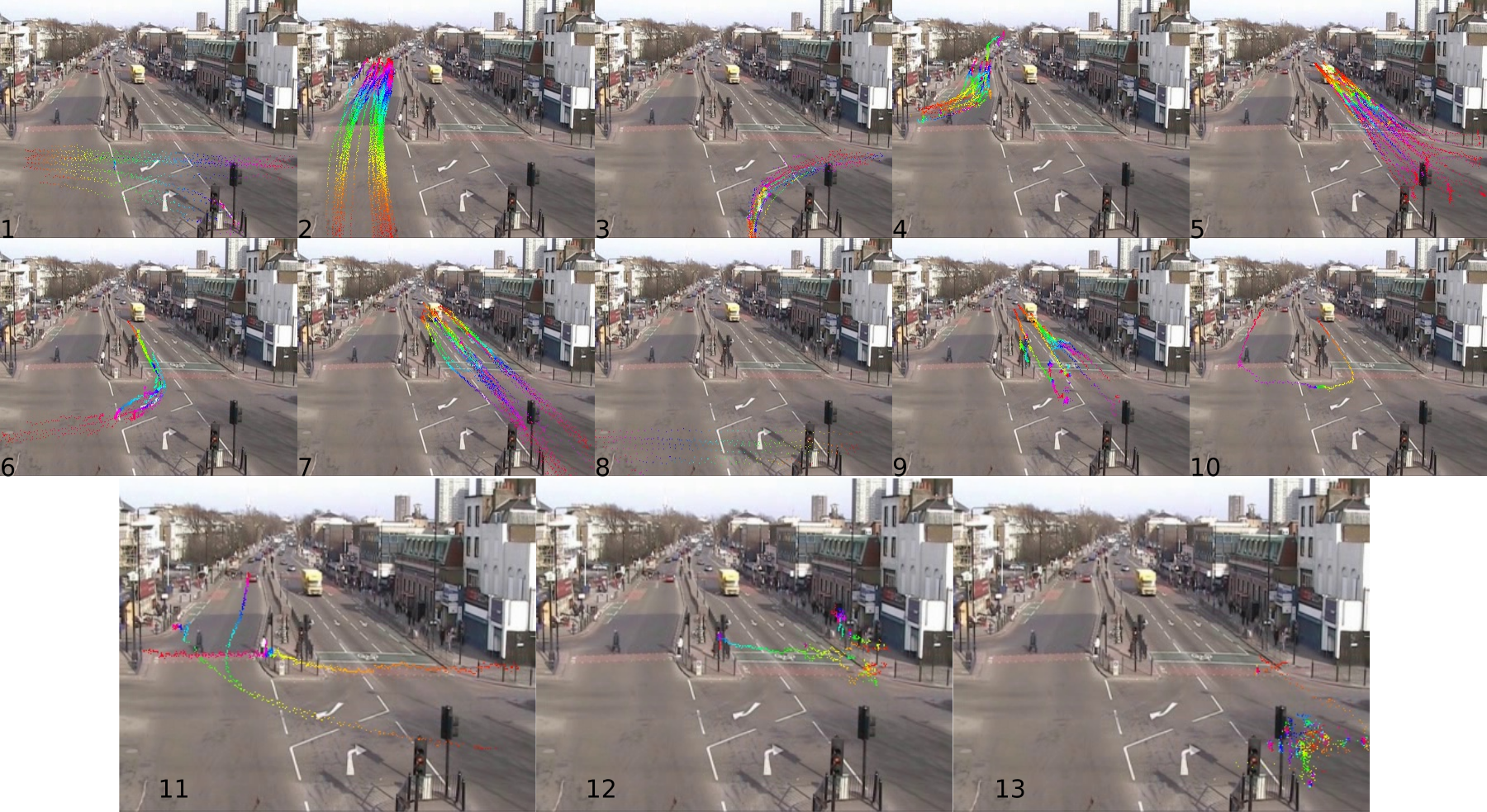}
\caption{Illustration of unsupervised clustering using mDPMM on QMUL trajectories. First two rows provide visual clue about the possible patterns in the scene. Last row images indicate rare patterns or possible outliers. The images labeled $5$, $7$, and $9$ can be grouped together to form a single class indicating the downward traffic flow.}
\label{fig_QMUL_CLUSTERS}
\end{figure}
\begin{figure}[h]
\centering
\includegraphics[width=0.48\textwidth]{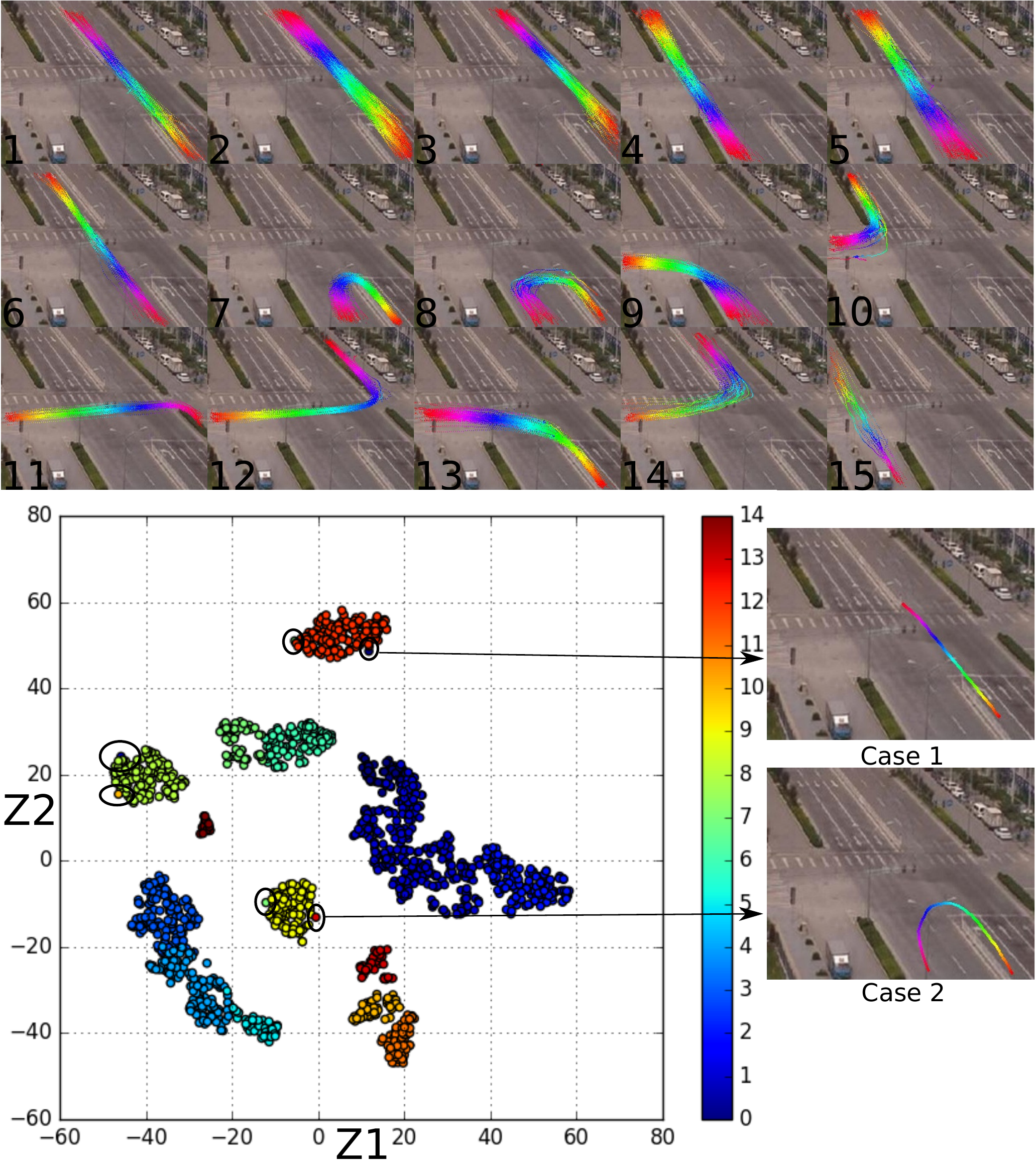}
\caption{Illustration of t-SNE on T15. $z1$ and $z2$ represent the dimensions after feature transformation. Top three rows represent trajectories of 15  classes. Most of the anomalies can be identified from the t-SNE visualization. For example, Case:1 the trajectory is anomalous due to the truncation and Case 2 is a normal trajectory. The latter is away from the respective class distribution due to the U-turn variation of the vehicle. }
\label{fig_T15_tSNE}
\end{figure}
The annotation aspects of unlabeled trajectories using mDPMM are shown in Fig.\ref{fig_QMUL_CLUSTERS}. Trajectory details are presented in Table~\ref{Tab:Dataset}. Since T15 dataset readily comes with associated class annotation, unsupervised clustering has not been not applied on this dataset.  Fig.\ref{fig_T15_tSNE} presents the t-SNE guided refinement.
\subsection{Experiments on Classification and Comparisons}
\begin{figure}[h]
\centering
{\includegraphics[width=0.46\textwidth]{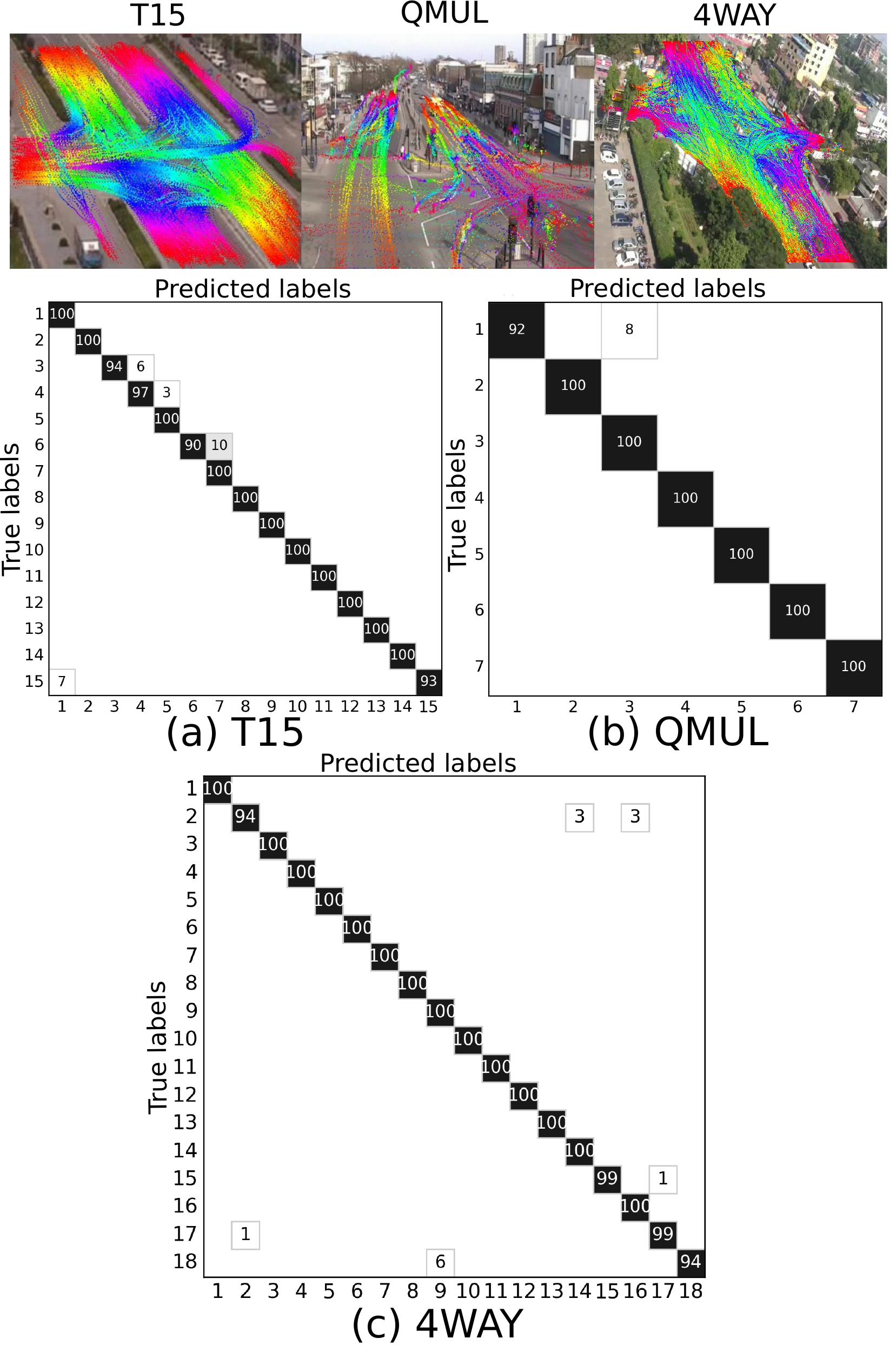}}
\caption{Illustration of test results using normalized confusion matrices (in \%) for T15, QMUL and 4WAY datasets.}
\label{fig_Classification}
\end{figure}
Trajectories of T15, QMUL and 4WAY datasets have been used for classification.  Classification results are  shown in Fig.~\ref{fig_Classification} and summarized in Table~\ref{Tab:Accuracy}. It can be observed that the proposed method performs accurate classification across all datasets. We have randomly selected $75$\% of the trajectories for training and the rest for testing.
Our proposed classification method  has been compared with other state-of-the-art classification methods such as HAR-CNN\cite{yang2015deep}, LSTM\cite{AKhosroshahi}, LSTM+CNN \cite{wu2015modeling} typically used for time series data classification. We have converted the input trajectories to  $128$ samples by downsampling or upsampling depending on their size. The comparative results are shown in Table~\ref{Tab:Accuracy}. The results reveal, our proposed method performs better than the existing approaches across all datasets. However, classification without color gradient degrades slightly, even though it performs  better than most of the existing work. 
\begin{table}[h]
\caption {Comparison of classification accuracies on three datasets} 
\label{Tab:Accuracy} 
\begin{center}
    \begin{tabular}{| l | c | c | c |}
    \hline
    Method & T15  & QMUL & 4WAY \\ \hline\hline
    Proposed & \textbf{99.0}\% & \textbf{97.3}\% & \textbf{99.5}\% \\ 
    HAR-CNN \cite{yang2015deep} & 94.9\% & \textbf{97.3}\% & 98.7\% \\ 
    LSTM \cite{AKhosroshahi}) & 93.4\% & 88.6\% & 93.0\% \\ 
    LSTM+CNN \cite{wu2015modeling} & 93.4\% & 91.3\% & 94.1\%\\   
    Proposed (no gradient)  & 98.0\% & 94.6\% & 99.1\%\\    
    \hline
    \end{tabular}
\end{center}
\end{table} 
\subsection{Experiments on Anomaly Detection}
\begin{figure}[h]
  \centering
  \begin{minipage}[h]{0.48\textwidth}  
\includegraphics[scale=0.183]{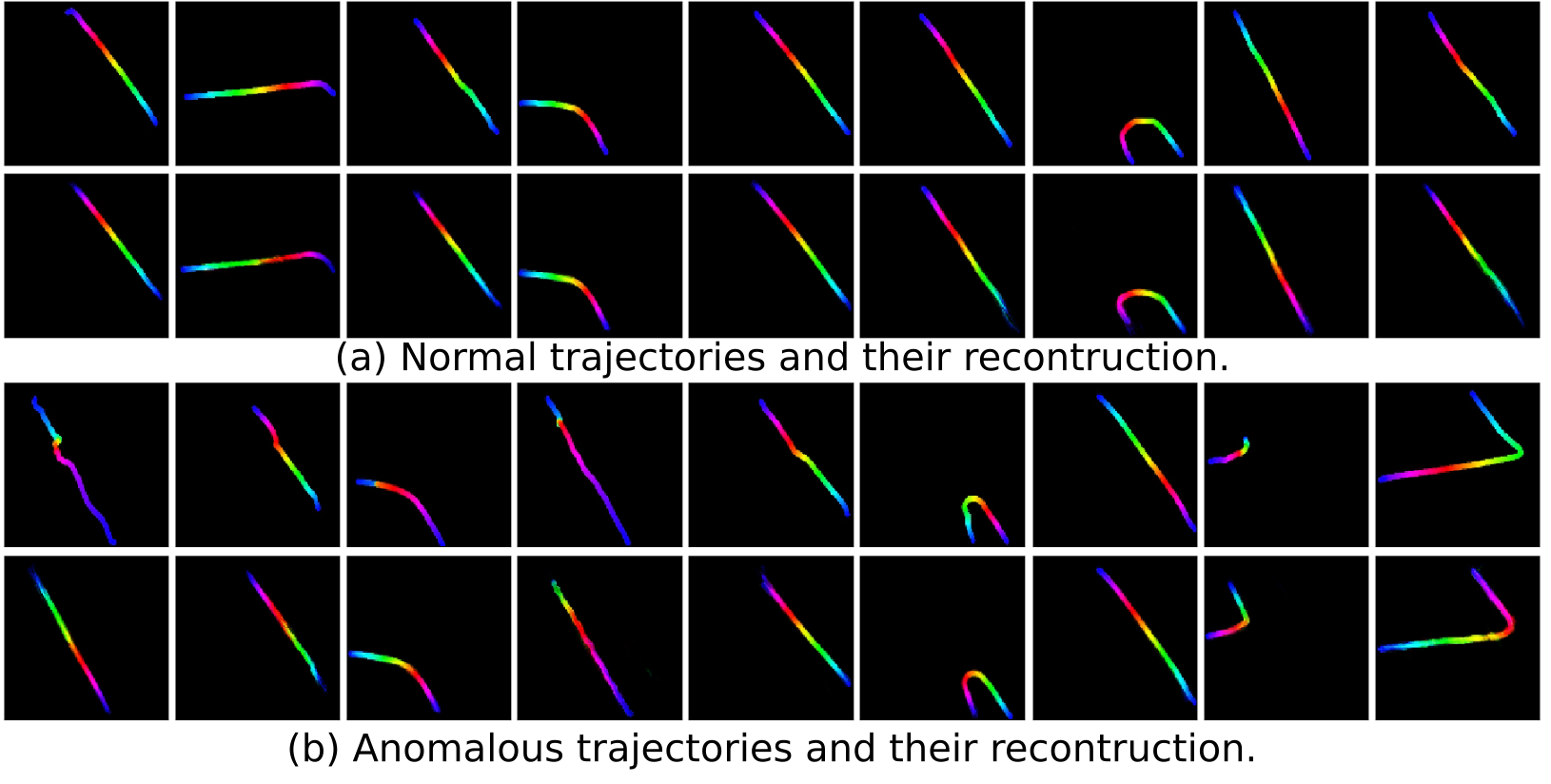}
\caption{Illustration of anomaly detection on T15 dataset. First row presents test trajectories and second row  presents corresponding reconstructed patterns. (a) Correct reconstruction happens for the non-anomalous trajectories. (b) Reconstruction fails on anomalous trajectories. Columns 2 and 5 represent lane change anomaly. Columns 1 and 3 represent speed variations. Column 4 represents vehicle stopping then moving and column 8 represents terminated trajectory. Columns 6, 7 and 9 represent vehicle moving in opposite direction of normal traffic.}
\label{fig_T15}  
  \end{minipage}  
  \begin{minipage}[h]{0.48\textwidth}
\subfigure[Reconstruction loss]{\includegraphics[scale=0.23]{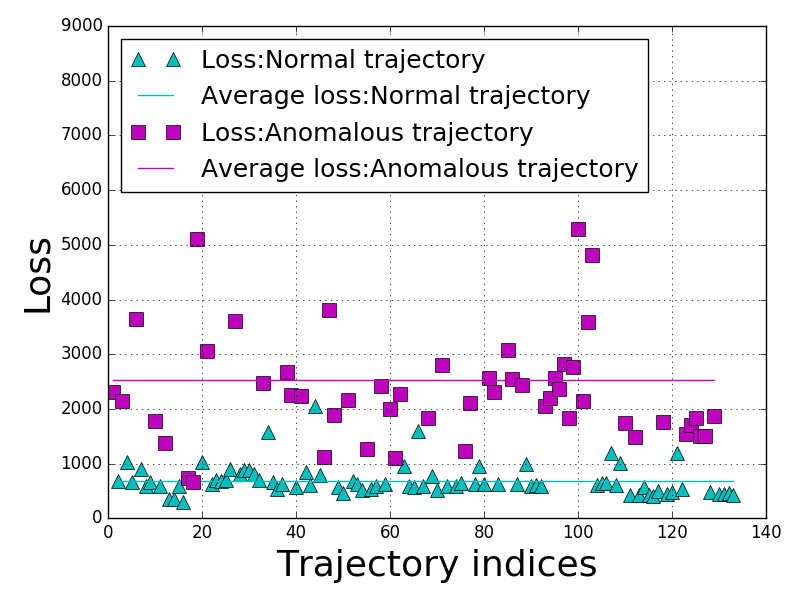}}
\subfigure[Confusion Matrix]{
\includegraphics[scale=0.25]{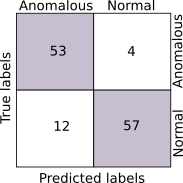}
}  
    \caption{ Depiction of anomaly detection. (a) Reconstruction loss ($l_i$) for normal and anomalous trajectories. The loss depends on the amount of deviation from the normal path. (b) Confusion matrix for anomaly detection experiments on T15 dataset.}
  \label{fig_LOSS_CURVE}    
  \end{minipage}
\end{figure}
\begin{table}[h]
\caption {Comparisons of anomaly detection} 
\label{Tab:Anomaly_Accuracy} 
\begin{center}
    \begin{tabular}{| l | c | c | c |}
    \hline
    Method & Accuracy  & Precision & Recall \\ \hline\hline
    Without Gradient & 49.2\% & 46.7\% & 85.9\% \\ 
    Without t-SNE  & 86.5\% & \textbf{83.1}\% & 87.5\% \\ 
    With t-SNE  & \textbf{87.3}\% & 81.5\% & \textbf{93.0}\% \\ 
    \hline
    \end{tabular}
\end{center}
\end{table}  
T15 dataset has been used to evaluate the anomaly detection framework. Reconstructions using VAE are depicted in Fig.~\ref{fig_T15}. Four kinds of anomalous trajectories are used in our experiments: (i) Trajectories terminating abruptly. (ii) Speed variation as compared to normal trajectories of the same class. (iii) Trajectories of objects traveling in opposite direction of the normal traffic. (iv) Trajectories corresponding to vehicles violating lane driving.  Since T15 dataset does not contain type three anomalous trajectories, we have created a few such trajectories by gradient conversion in reverse order. We have used two times the converged loss value as a threshold for detecting anomaly based on the empirical study on anomalous and normal trajectories as shown in Fig~\ref{fig_LOSS_CURVE}(a). Anomaly detection results are shown in Fig.~\ref{fig_LOSS_CURVE}(b). We have used 69 randomly selected normal trajectories that are not used in the training and 31 identified anomalous trajectories along with synthetically created ones. We have created $11$ synthetic trajectories for lane change and $15$ corresponding to each class for opposite direction driving anomalies. The comparisons of trajectory projection on image plane  using VAE under different conditions are presented in Table~\ref{Tab:Anomaly_Accuracy}. We are able to detect anomalies with an accuracy of 87.3\% when t-SNE is used. This reveals, without gradient representation, anomaly detection accuracy drops significantly (49.2\%).
\subsection{Comparison of Anomaly Detections}
Since video trajectory-based anomaly detection method using DNNs proposed in this paper is of the first kind, we could not find benchmark datasets that can be used in comparison with neural network-based anomaly detection. Hence we have performed high-level comparison with the state-of-the-art anomaly detection techniques presented in~\cite{2013_ICCV_AbnormalEvent} and~\cite{2017_ISNN_Spatio_Temp_AE} using the input reconstruction property. The work proposed in \cite{2013_ICCV_AbnormalEvent} uses sparse combination learning for learning normal behavior, while \cite{2017_ISNN_Spatio_Temp_AE} learns the model from the spatio-temporal video segments using Autoencoder. Several experiments have been conducted on QMUL dataset. Training videos have been created by splitting the original traffic video into $42$ segments starting from the frame number $8610$ by eliminating anomalous segments from the scene. Testing has been conducted using the video segment prior to the frame number $8610$. We have trained our proposed architecture using trajectories obtained with the help of the method proposed in~\cite{zhou2013measuring} with $\delta = 836$ for the testing. Training for the method proposed in \cite{2013_ICCV_AbnormalEvent} has been done using the same configuration as reported in their work, while testing has been conducted with an error threshold of $0.4$. For training the model proposed in \cite{2017_ISNN_Spatio_Temp_AE}, we have used a sequence length ($T$) = 10, batch size = 4 and number of epochs = 200. 

\begin{figure}[h]
\centering
{\includegraphics[width=0.49\textwidth]{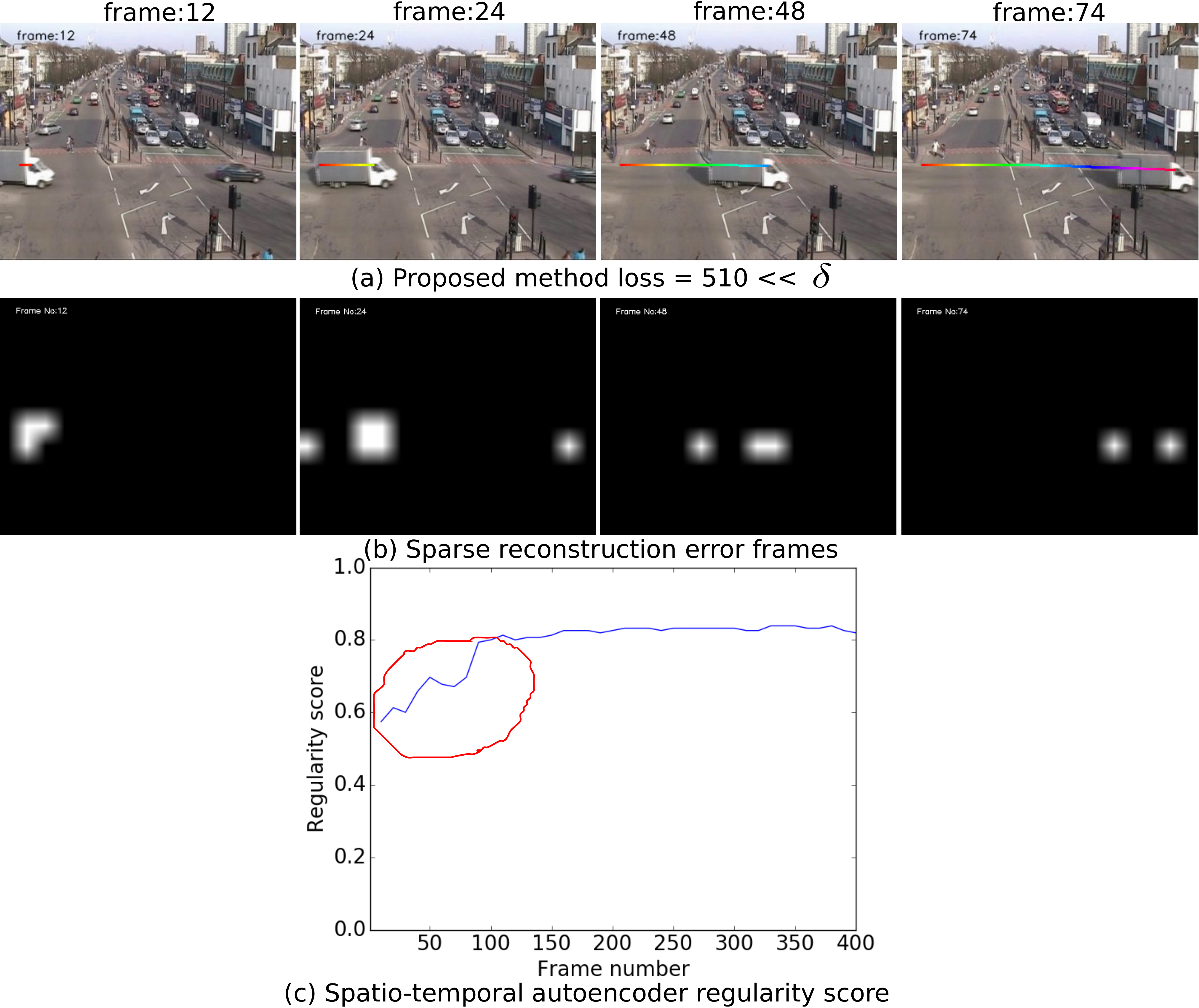}}
\caption{Illustration of false alarms in the sparse reconstruction technique \cite{2013_ICCV_AbnormalEvent} and spatio-temporal Autoencoder-based method\cite{2017_ISNN_Spatio_Temp_AE}. Though the traffic is open for vehicles at the junction for east and south bound traffic from top-peft lane, anomalies are reported from the scene for these methods. (a) A trajectory corresponding to a truck in color gradient form in different frames. (b) The respective sparse reconstruction error frames using \cite{2013_ICCV_AbnormalEvent}. White patches represent the anomalies detected by the method proposed in \cite{2013_ICCV_AbnormalEvent}. It can be seen that several false positives are present throughout the sequence. (c) The regularity score for the scene using \cite{2017_ISNN_Spatio_Temp_AE} shown for the frame sequence. The highlighted portion indicates some anomaly. }
\label{fig_Anomaly_False_Alarm1}
\end{figure}

\begin{figure}[h]
\centering
{\includegraphics[width=0.49\textwidth]{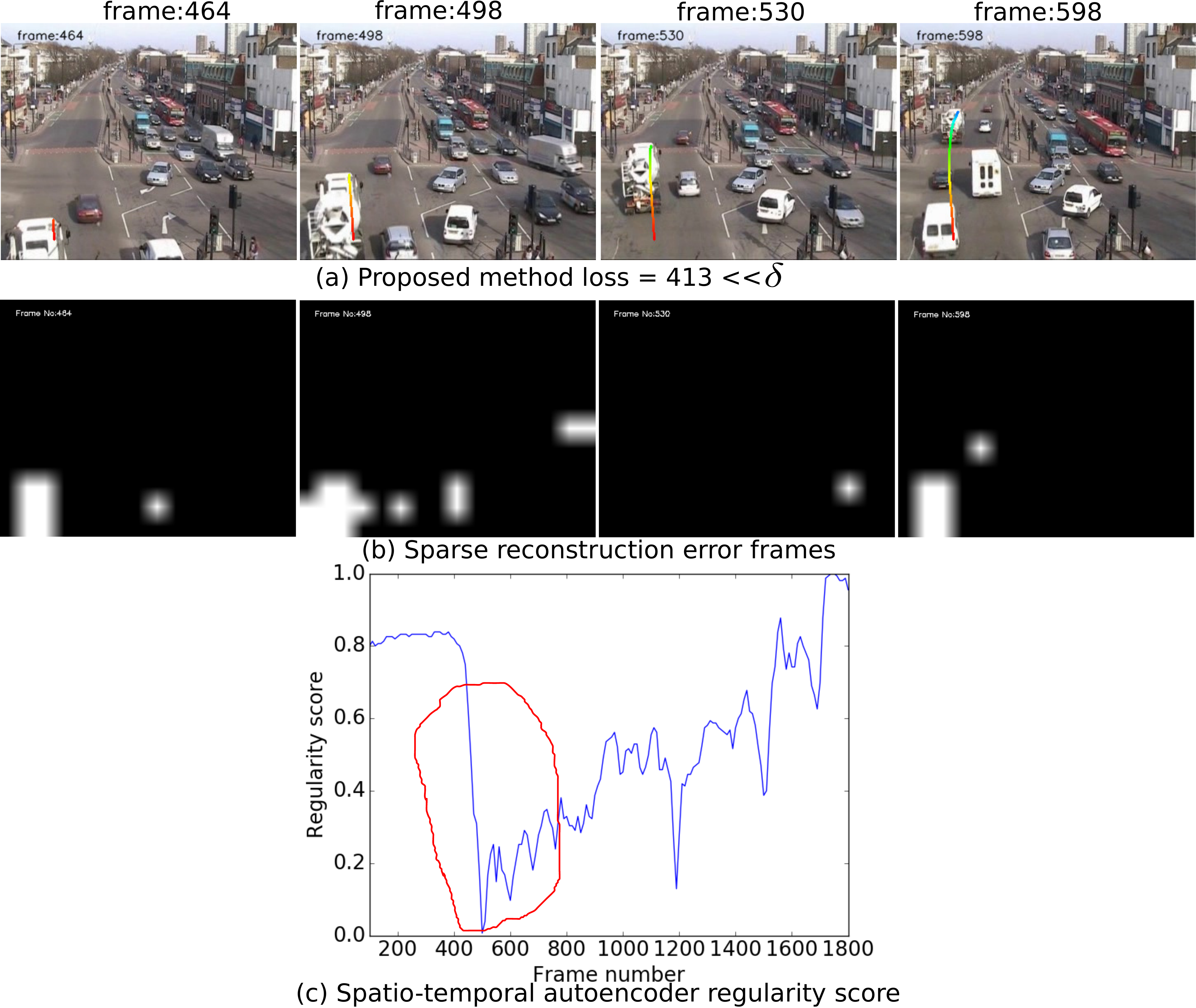}}
\caption{Depiction of anomaly detection for a scene, where the signal opens for north-bound and south-bound traffics. Though the traffic movement seems to be normal, false positives are reported using \cite{2013_ICCV_AbnormalEvent} and \cite{2017_ISNN_Spatio_Temp_AE}. (a) A trajectory corresponding to a truck in color gradient form in different frames. (b) Corresponding sparse reconstruction error frames using \cite{2013_ICCV_AbnormalEvent}. False positives are present even for other vehicles. (c) The regularity score for the scene using \cite{2017_ISNN_Spatio_Temp_AE} during the frame sequence. False positives can be seen for longer duration when heavy traffic flow is underway during green signal.}
\label{fig_Anomaly_False_Alarm2}
\end{figure}

\begin{figure*}[h]
\centering
{\includegraphics[width=0.98\textwidth]{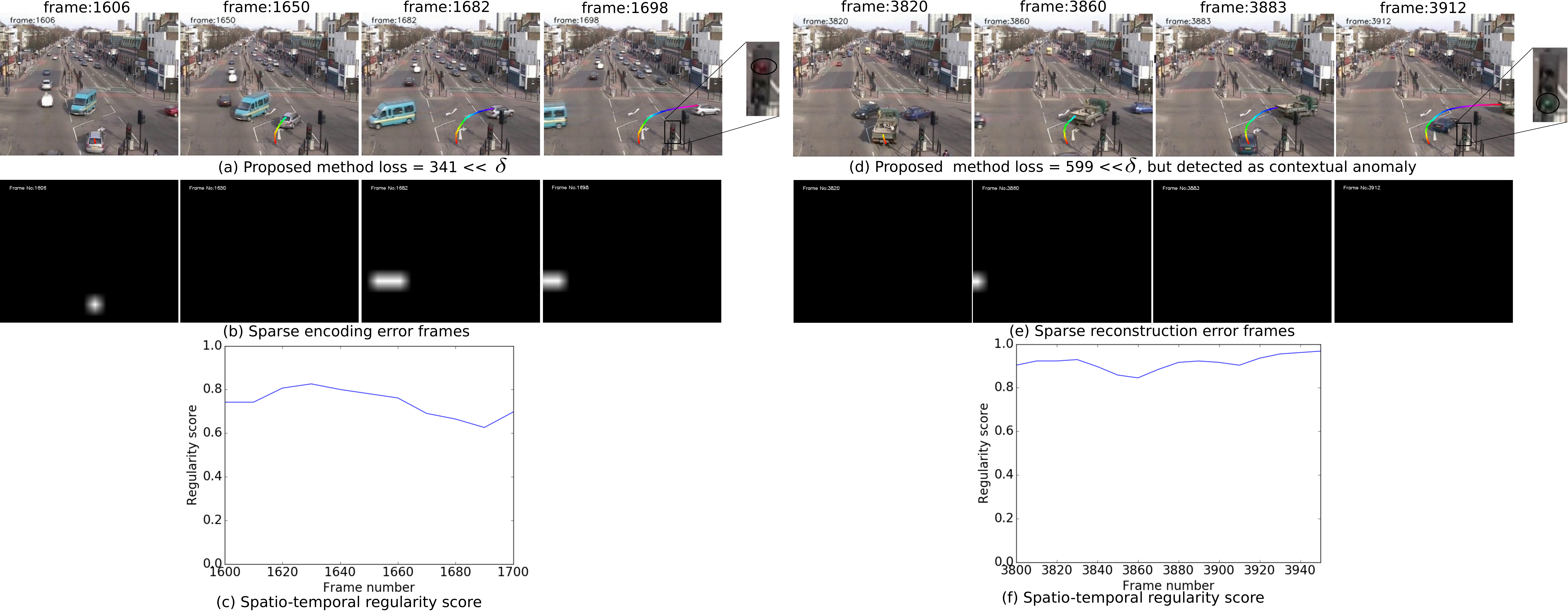}}
\caption{Depiction of contextual anomaly detection. When the traffic signal is green, only north-bound and south-bound traffics are allowed. Going right or left is allowed only after the end of the signal. Traffic scenario depicted through (a-c) is a normal condition, while (d-f) depict contextual anomaly. (a) A trajectory corresponding to a car in color gradient during normal traffic flow. When the signal is red, the waiting vehicles are allowed to go to the right or left, i.e. the pattern is an allowed one. The loss value is well below the anomaly threshold, indicating that it is a normal flow. (b) The respective sparse reconstruction error frames  using \cite{2013_ICCV_AbnormalEvent}. Though false positives are not observed throughout for the tracked vehicle during this period, it is present in some frames. False positives can also be observed for the vehicle turning left. (c) The regularity score for the scene using \cite{2017_ISNN_Spatio_Temp_AE}. The regularity score does not indicate any anomaly. (d) A trajectory corresponding to a truck in color gradient form during an anomalous traffic flow. As the signal is green, even though the no vehicles can be seen heading south ward, vehicles are not supposed to cross towards east side. Though the loss is less than the threshold, this is categorized as an unknown anomaly using our method. (e)  Corresponding sparse reconstruction error using \cite{2013_ICCV_AbnormalEvent} with no anomalies detected. (f) The regularity score for the scene using \cite{2017_ISNN_Spatio_Temp_AE}. The regularity score does not indicate any anomaly, though there is a contextual anomaly.}
\label{fig_Anomaly_contextual}
\end{figure*}

\begin{figure}[h]
\centering
{\includegraphics[width=0.49\textwidth]{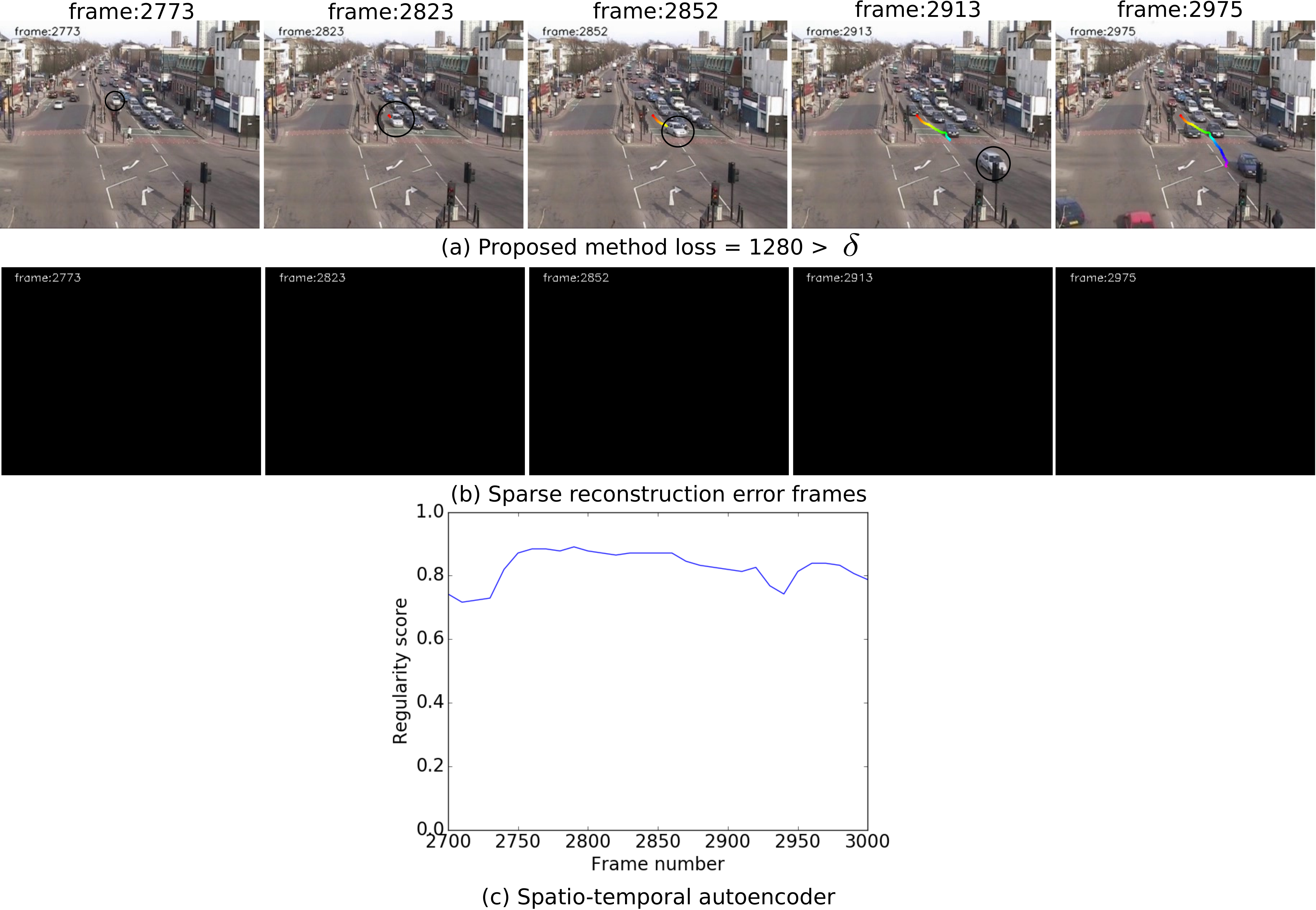}}
\caption{Illustration of lane change anomaly with truncated trajectory using three different methods. (a) The highlighted vehicle gets tracked very late and the tracking fails and wrong trajectories are created. However, our method can detect it as an anomaly. (b) Corresponding  sparse reconstruction  error using \cite{2013_ICCV_AbnormalEvent} with no possible anomaly. (c) The regularity score of the scene using \cite{2017_ISNN_Spatio_Temp_AE}. The regularity score does not indicate any anomaly.}
\label{fig_Actual_anomaly}
\end{figure}

The test results are depicted through Figs.~\ref{fig_Anomaly_False_Alarm1}-\ref{fig_Actual_anomaly}. It can be observed that both the methods proposed in~\cite{2013_ICCV_AbnormalEvent} and ~\cite{2017_ISNN_Spatio_Temp_AE} report several false positives on the QMUL dataset. Moreover, these methods cannot detect contextual anomalies. A deeper analysis reveals that the false positives are mainly due to the unseen characteristics present in the scene with heterogeneous data, making it difficult to learn all spatio-temporal features. Such methods can work only when the video duration is long enough that can learn all types of object motions possible within a scene. However, it may be difficult to train as separating  normal video segments from the anomalous can be very challenging when anomalies are present throughout the video. As our method is trajectory-based, individual trajectories can be characterized as normal or abnormal rather than declaring a video segment normal or anomalous. Moreover, training a deep neural network using video frames can be time consuming. On the contrary, a trajectory has been condensed into a single video frame as done in our method. In a nutshell, we are using the advantages of conventional trajectory extraction methods as well as the feature extraction capabilities of deep neural network to achieve classification which is fast.  Table~\ref{Tab:Anomaly_Comparisons} summarizes the comparative results.   


\begin{table}[h]
\caption {Comparisons of anomaly detection with State-of-art} 
\label{Tab:Anomaly_Comparisons} 
\begin{center}
    \begin{tabular}{| p{2.5cm} | p{1.5cm} | p{1.5cm} | p{1.5cm} |}
    \hline
    \textbf{Parameters} & \textbf{Proposed method}  & \textbf{Sparse reconstruction}\cite{2013_ICCV_AbnormalEvent} & \textbf{Spatio-temporal autoencoder}\cite{ 2017_ISNN_Spatio_Temp_AE} \\ \hline\hline
    False alarm rate & Low & High & High\\ \hline
    Unknown anomaly detection  & Yes & Yes & Yes \\ \hline
    Contextual anomaly detection & Yes & No & No \\ \hline
    Training difficulty  & Low & High & High \\ \hline   
    Anomaly localization  & Yes & Yes & No \\ \hline   
    Detection time  & Once trajectory is available & Per frame & Per sequence length \\            
    \hline
    \end{tabular}
\end{center}
\end{table}  

\subsection{Discussions and Limitations}
Key to accurate anomaly detection lies in training the model with normal trajectories. Apart from mDPMM-based clustering, t-SNE visualization plays an important role in eliminating anomalous trajectories. The need for a classifier is to detect known anomalies such as traffic rule violations by vehicles. While unknown anomalies are detected using VAE, the CNN classifier helps to identify known anomalies and to localize the path of unknown anomalies. The loss values in terms of KLD and likelihood are justified as they represent the distance of the trajectories from the allowed class distributions.  A small offset from the converged loss can be a good estimate of the threshold. CNN classifier performs with higher accuracy as compared to other methods. 

Some of the limitations of the proposed method are: (i) The method is tracking dependent. However, with improved tracking, we can overcome this issue. (ii) A large number of training samples need to be available to learn the allowed paths in a traffic junction.

\section{Conclusions}
\label{sec:Conclusions}
The key idea behind this work is to represent time varying visual data using color gradient form in order to train DNN-based systems for encoding temporal features. This method combines traditional object tracking-based results to be combined with neural network-based methods to use the advantages of both systems. It has been observed through experiments that the proposed color gradient feature using CNN performs better than existing classifiers. We are also able to detect a few types of trajectory anomalies using the proposed architecture. It performs better than some of the existing reconstruction-based anomaly detection methods. We plan to extend this work to develop a real-time anomaly detection system for traffic intersections using online trajectories which will be able to detect discussed anomalies as well as other anomalies such as over-speeding. We also plan to explore this method for time series data analysis in other domains.


%

%
%
%
\section*{Acknowledgment}
We gratefully acknowledge the support of NVIDIA Corporation with the donation of the Quadro P5000 GPU used for this research.


\ifCLASSOPTIONcaptionsoff
  \newpage
\fi



%
\bibliographystyle{plain}

%

\begin{IEEEbiography}[{\includegraphics[width=1in,height=1.25in,clip,keepaspectratio]{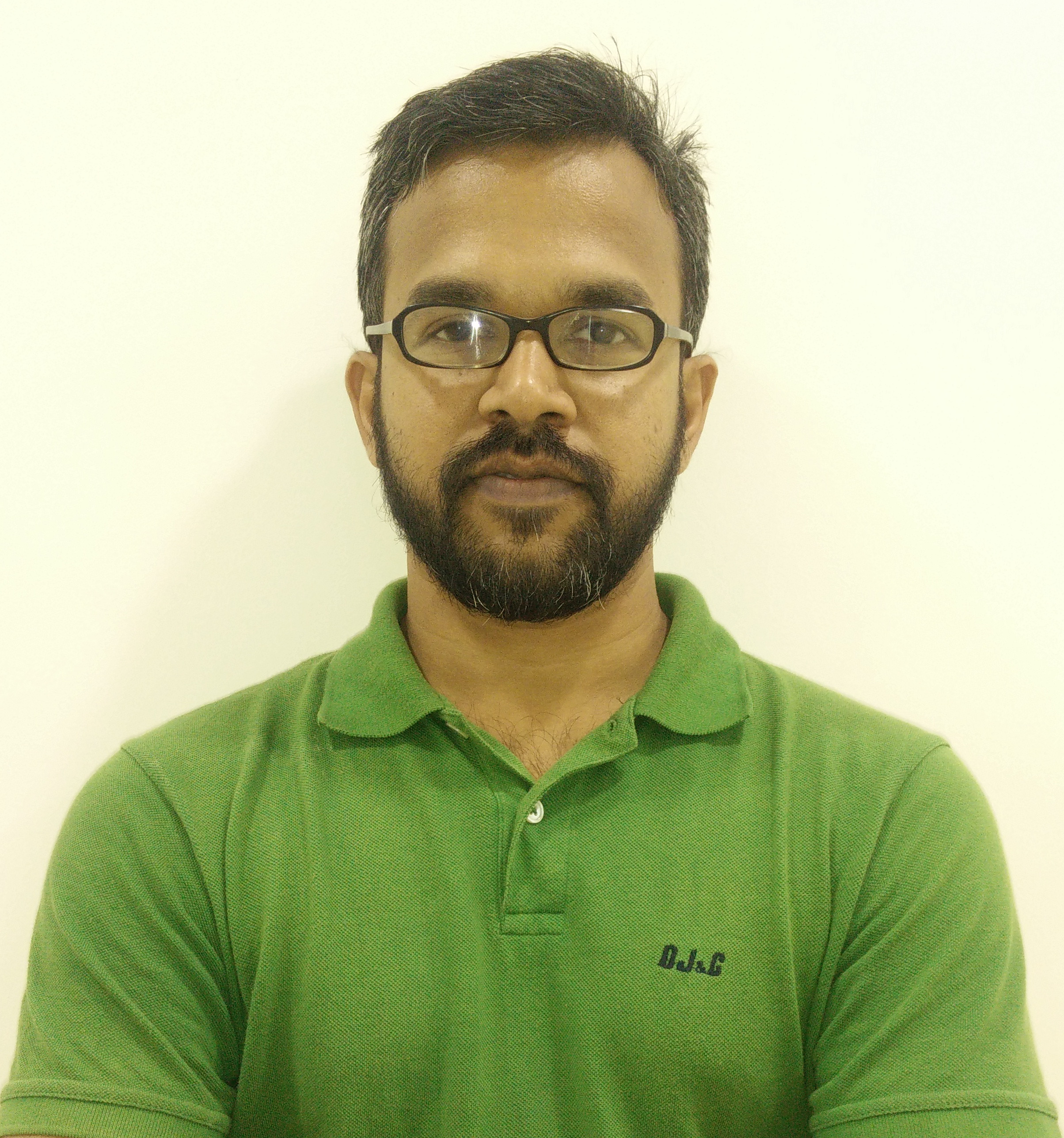}}]{Kelathodi Kumaran Santhosh} is a research scholar in the School of Electrical Sciences, IIT Bhubaneswar, India. He joined a Ph.D. program for resuming his research work that can help humanity. His interests are in the development of vision based applications that can replace human factor. He is a member of IEEE. Prior to joining IIT Bhubaneswar, he worked for Huawei Technologies India Pvt. Ltd. for 10 years (2005-2015) and in Defence Research Development Organization (DRDO) as a Scientist for around 2 years (2003-2004). During his tenure with Huawei, he has worked in many signalling protocols such as Diameter, Radius, SIP etc. in the role of a developer, technical leader, project manager and also served the product lines HSS, CSCF etc. in Huawei China as a support engineer for closer to 1.5 years. In DRDO, he worked in the field of object tracking algorithms based on the data received from radars. More information on Santhosh can be found at https://sites.google.com/site/santhoshkelathodi.
\end{IEEEbiography}
\begin{IEEEbiography}[{\includegraphics[width=1in,height=1.25in,clip,keepaspectratio]{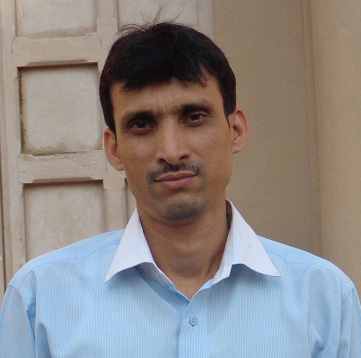}}]{Dr. Debi Prosad Dogra} is an Assistant Professor in the School of Electrical Sciences, IIT Bhubaneswar, India. He received his M.Tech degree from IIT Kanpur in 2003 after completing his B.Tech. (2001) from HIT Haldia, India. After finishing his masters, he joined Haldia Institute of Technology as a faculty members in the Department of Computer Sc. \& Engineering (2003-2006). He has worked with ETRI, South Korea during 2006-2007 as a researcher. Dr. Dogra has published more than 45 international journal and conference papers in the areas of computer vision, image segmentation, and healthcare analysis.  He is a member of IEEE. More information on Dr. Dogra can be found at \url{http://www.iitbbs.ac.in/profile.php/dpdogra}.
\end{IEEEbiography}
\begin{IEEEbiography}[{\includegraphics[width=1in,height=1.25in,clip,keepaspectratio]{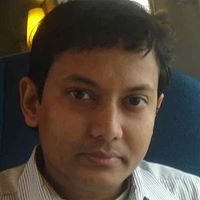}}]{Dr. Partha Pratim Roy} has obtained his M.S. and Ph. D. degrees in the year of 2006 and 2010, respectively at Autonomous University of Barcelona, Spainis. Presently he is an Assistant Professor in the Department of Computer Science and Engineering, IIT Roorkee, India in 2014. Prior to joining, IIT Roorkee, Dr. Roy was with Advanced Technology Group, Samsung Research Institute Noida, India during 2013-2014. Dr. Roy was with Synchromedia Lab, Canada in 2013 and RFAI Lab, France in 2012 as postdoctoral research fellow. His research interests are Pattern Recognition, Multilingual Text Recognition, Biometrics, Computer Vision, Image Segmentation, Machine Learning, and Sequence Classification. He has published more than 65 papers in international journals and conferences. 
\end{IEEEbiography}

\begin{IEEEbiography}[{\includegraphics[width=1in,height=1.25in,clip,keepaspectratio]{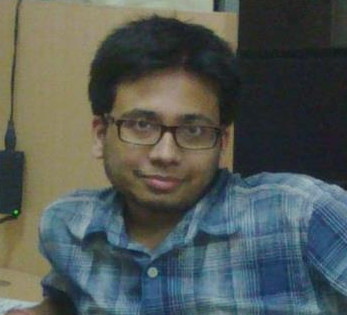}}]{Dr. Adway Mitra} is a researcher, interested in Machine Learning and Data Mining, and especially in the application of these techniques to solve problems affecting the world. More specifically, he is interested in data-driven modeling and simulation of complex spatio-temporal processes. His background is in Computer Science and Engineering, and PhD thesis was related to semantic Video Analytics, using Bayesian modeling techniques. Many of the techniques and concepts he developed during PhD may be extended to spatio-temporal processes in other domains. By doing so, he intend to build a career in interdisciplinary research. He is currently focusing on Climate Informatics - application of Computer Science (especially Data Science) concepts to solve problems in Climate Science. He is particularly interested in the following questions in this domain:
1) Realistic simulation of climatic processes, through stochastic processes 2) Understanding and Modeling the dynamics of Indian Monsoon and its various vagaries like onset, withdrawal and active/break spells 3) Extreme events - their various statistical properties, and links between different extreme events 4) Identification of widespread and long-lasting anomalies such as droughts and heat waves in huge volumes of climatic data 5) Causal relationships between different events, aimed at attribution. 
\end{IEEEbiography}




\end{document}